\documentclass[runningheads]{llncs}
\usepackage{graphicx}
\usepackage{amsmath,amssymb} 
\usepackage{color}
\usepackage{hyperref}
\hypersetup{
	colorlinks=false
}

\usepackage{booktabs}

\usepackage[linesnumbered,ruled]{algorithm2e}

\newcommand{\prg}[1]{\vspace{0.1cm}\noindent\textbf{#1}~~}

\newcommand{\bX}{\mathbf{X}}
\newcommand{\bx}{\mathbf{x}}
\newcommand{\bY}{\mathbf{Y}}
\newcommand{\bU}{\mathbf{U}}
\newcommand{\bW}{\mathbf{W}}
\newcommand{\bw}{\mathbf{w}}
\newcommand{\bz}{\mathbf{z}}
\def\bmu{\boldsymbol\mu}
\newcommand{\btX}{\mathbf{\tilde{X}}}

\newcommand{\phantomone}{\phantom{a}}
\newcommand{\phantomtwo}{\phantom{aa}}

\begin{document}
	\title{Statistically-motivated Second-order Pooling\thanks{This work was supported in part by the Swiss National Science Foundation.}} 
	
	\titlerunning{Statistically-motivated Second-order Pooling}
	\author{Kaicheng Yu\orcidID{0000-0002-0186-3399}  \and
		Mathieu Salzmann\orcidID{0000-0002-8347-8637} }
	\authorrunning{K. Yu and M. Salzmann}
	
	\institute{CVLab, EPFL, 1015 Lausanne, Switzerland
		\email{\{kaicheng.yu,mathieu.salzmann\}@epfl.ch}}
	
	\maketitle              


\begin{abstract}

Second-order pooling, a.k.a.~bilinear pooling, has proven effective for deep learning based visual recognition. However, the resulting second-order networks yield a final representation that is orders of magnitude larger than that of standard, first-order ones, making them memory-intensive and cumbersome to deploy. Here, we introduce a general, parametric compression strategy that can produce more compact representations than existing compression techniques, yet outperform both compressed and uncompressed second-order models. Our approach is motivated by a statistical analysis of the network's activations, relying on operations that lead to a Gaussian-distributed final representation, as inherently used by first-order deep networks.
As evidenced by our experiments, this lets us outperform the state-of-the-art first-order and second-order models on several benchmark recognition datasets.

\end{abstract}


\section{Introduction}

Visual recognition is one of the fundamental goals of computer vision. Over the years, second-order representations, i.e., region covariance descriptors, have proven more effective than their first-order counterparts~\cite{Arandjelovic13,Dalal05,Lazebnik06,Perronnin10} for many tasks, such as pedestrian detection~\cite{Tuzel07}, material recognition~\cite{Cimpoi14} and semantic segmentation~\cite{Carreira12}. 
More recently, convolutional neural networks (CNNs) have achieved unprecedented performance in a wide range of image classification problems~\cite{Krizhevsky12,He16,Huang17a}. Inspired by the past developments in handcrafted features, several works have proposed to replace the fully-connected layers with second-order pooling strategies, essentially utilizing covariance descriptors within CNNs~\cite{Lin15,Ionescu15,Li17a,Lin17a}. This has led to second-order or bilinear CNNs whose representation power surpasses that of standard, first-order ones.
	
One drawback of these second-order CNNs is that vectorizing the covariance descriptor to pass it to the classification layer, as done in~\cite{Lin15,Ionescu15,Li17a,Lin17a}, yields a vector representation that is orders of magnitude larger than that of first-order CNNs, thus making these networks memory-intensive and subject to overfitting. While compression strategies have been proposed~\cite{Gao16,Kong17}, they are either nonparametric~\cite{Gao16}, thus limiting the representation power of the network, or designed for a specific classification formalism~\cite{Kong17}, thus restricting their applicability.

In this paper, we introduce a general, parametric compression strategy for second-order CNNs. As evidenced by our results, our strategy can produce more compact representations than~\cite{Gao16,Kong17}, with as little as 10\% of their parameters, yet significantly outperforming these methods, as well as the state-of-the-art first-order~\cite{He16,Simonyan15} and uncompressed second-order pooling strategies~\cite{Lin15,Ionescu15,Li17a,Lin17a}.

\begin{figure}[t]
	\begin{center}
		\includegraphics[width=0.85\linewidth]{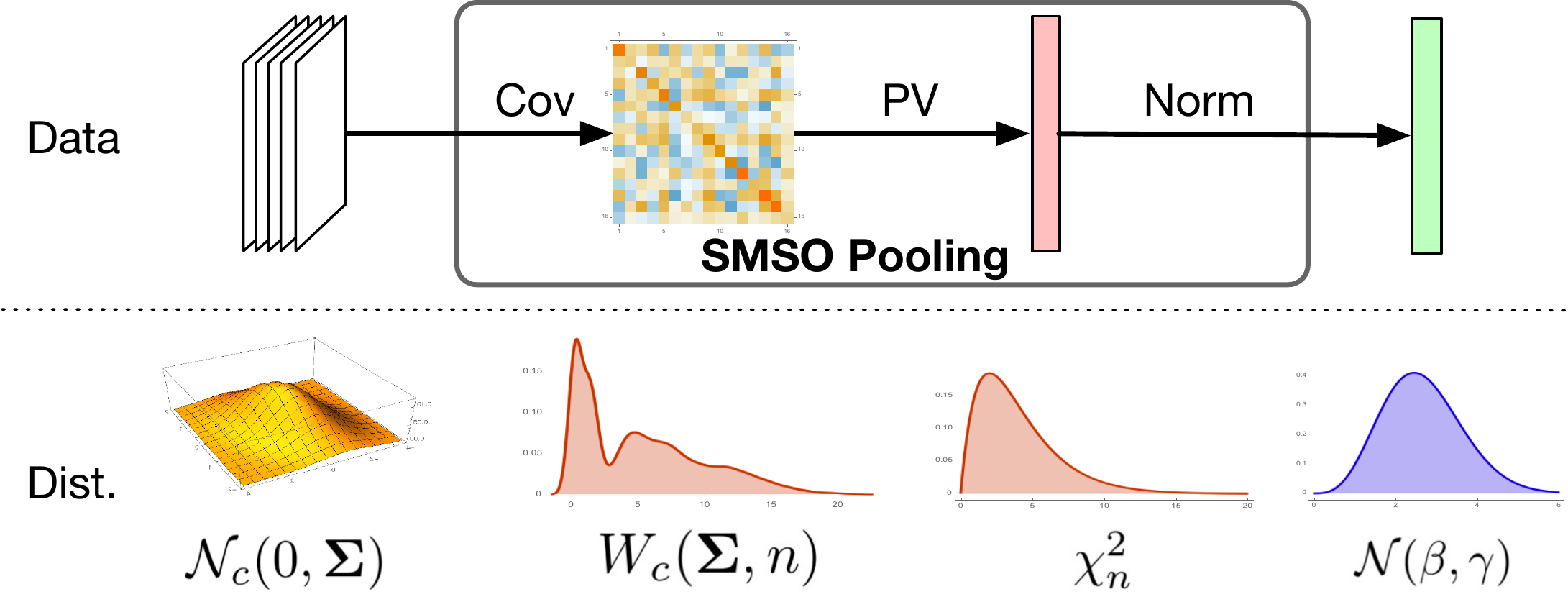}
	\end{center}
	\vspace{-0.5cm}
	\caption{{\bf Statistically-Motivated Second-Order (SMSO) pooling.} {\bf Top:} Our parametric compression strategy vectorizes a covariance matrix and normalizes the resulting vector. {\bf Bottom:} Each of these operations yields a well-defined distribution of the data, thus resulting in a consistent framework, whose final representation follows a Gaussian distribution, as state-of-the-art first-order deep networks.}  
	\label{fig:smso}
\end{figure}

Unlike most deep learning architectures, our approach is motivated by a statistical analysis of the network's activations. In particular, we build upon the observation that 
first-order networks inherently exploit Gaussian distributions
for their feature representations. This is due to the fact that, as discussed in~\cite{Goodfellow16,Ioffe15} and explained by the Central Limit Theorem, the outputs of \textit{linear} layers, and thus of operations such as global average pooling, follow a multivariate Gaussian distribution.
The empirical success of such Gaussian distributions of feature representations in first-order deep networks motivated us to design a compression strategy such that the final representation also satisfies this property.

To this end, as illustrated by Fig.~\ref{fig:smso}, we exploit the fact that the covariance matrices resulting from second-order pooling follow a Wishart distribution~\cite{Johnson14}. We then introduce a parametric vectorization (PV) layer, which compresses the second-order information while increasing the model capacity by relying on trainable parameters.
We show that our PV layer outputs a vector whose elements follow $\chi^2$ distributions, which motivates the use of a square-root normalization that makes the distribution of the resulting representation converge to a Gaussian, as verified empirically in Section~\ref{subsec:empirical_dist}.
These operations rely on basic algebraic transformations, and can thus be easily integrated into any deep architecture and optimized with standard backpropagation.

We demonstrate the benefits of our statistically-motivated second-order (SMSO) pooling strategy on standard benchmark datasets for second-order models, including the Describing Texture Dataset (DTD)~\cite{Cimpoi14}, the Material in Context (MINC) dataset~\cite{Bell15} and the scene recognition MIT-Indoor dataset~\cite{Quattoni09}. Our approach consistently outperforms the state-of-the-art second-order pooling strategies, independently of the base network used (i.e., VGG-D~\cite{Simonyan15} or ResNet-50~\cite{He16}), as well as these base networks themselves. Our code is publicly available at \url{https://github.com/kcyu2014/smsop}.


\section{Related Work}
Visual recognition has a long history in computer vision. Here, we focus on the methods that, like us, make use of representations based on second-order information to tackle this task. In this context, the region covariance descriptors (RCDs) of~\cite{Tuzel07} constitute the first attempt at leveraging second-order information. Similarly, Fisher Vectors~\cite{Arandjelovic13} also effectively exploit second-order statistics. Following this success, several metrics have been proposed to compare RCDs~\cite{Arsigny06,Pennec06,Quang14,Sra12}, and they have been used in various classification frameworks, such as boosting~\cite{Freund97a}, kernel Support Vector Machines~\cite{Vapnik98}, sparse coding~\cite{Cherian14,Guo10,Wang16d} and dictionary learning~\cite{Sra11,Harandi12,Li13c,Harandi15}. In all these works, however, while the classifier was trained, no learning
was involved in the computation of the RCDs.

To the best of our knowledge,~\cite{Harandi14}, and its extension to the log-Euclidean metric~\cite{Huang15}, can be thought of as the first attempts to learn RCDs. This, however, was achieved by reducing the dimensionality of input RCDs via a single transformation, which has limited learning capacity. In~\cite{Huang17a}, the framework of~\cite{Harandi14} was extended to learning multiple transformations of input RCDs. Nevertheless, this approach still relied on RCDs as input.
The idea of incorporating second-order descriptors in a deep, end-to-end learning paradigm was introduced concurrently in~\cite{Ionescu15} and~\cite{Lin15}. The former introduced the DeepO\textsubscript{2}P operation, consisting of computing the covariance matrix of convolutional features. The latter proposed the slightly more general idea of bilinear pooling, which, in principle, can exploit inner products between the features of corresponding spatial locations from different layers in the network. 
In practice, however, the use of cross-layer bilinear features does not bring a significant boost in representation power~\cite{Gao16,Lin17a}, and bilinear pooling is therefore typically achieved by computing the inner products of the features within a single layer, thus becoming essentially equivalent to second-order pooling. 

A key to the success of second-order pooling is the normalization, or transformation, of the second-order representation. In~\cite{Ionescu15}, the matrix logarithm was employed, motivated by the fact that covariance matrices lie on a Riemannian manifold, and that this operation maps a matrix to its tangent space, thus producing a Euclidean representation. By contrast,~\cite{Lin15} was rather inspired by previous normalization strategies for handcrafted features~\cite{Arandjelovic13,Perronnin10}, 
and proposed to perform an element-wise square-root and $\ell_2$ normalization after vectorization of the matrix representation. 
More recently,~\cite{Li17a,Lin17a} introduced a matrix square-root normalization strategy that was shown to outperform the other transformation techniques. 

All the above-mentioned methods simply vectorize the second-order representation, i.e., covariance matrix. As such, they produce a final representation whose size scales quadratically with the number of channels in the last convolutional feature map, thus being typically orders of magnitude larger than the final representation of first-order CNNs. To reduce the resulting memory cost and parameter explosion,
several approaches have been proposed to compress second-order representations while preserving their discriminative power. The first attempt at compression was achieved by~\cite{Gao16}, which introduced two strategies, based on the idea of random projection, to map the covariance matrices to vectors. These projections, however, were not learned, thus not increasing the capacity of the network and producing at best the same accuracy as the bilinear CNN of~\cite{Lin15}. In~\cite{Kong17}, a parametric strategy was employed to reduce the dimensionality of the bilinear features. While effective, this strategy was specifically designed to be incorporated 
in a bilinear Support Vector Machine. 

By contrast, here, we introduce a parametric compression approach that can be incorporated into any standard deep learning framework. Furthermore, our strategy is statistically motivated so as to yield a final representation whose distribution is of the same type as that inherently used by first-order deep networks.
As evidenced by our experiments, our method can produce more compact representations than existing compression techniques, yet outperforms the state-of-the-art first-order and second-order models.

Note that higher-order information has also been exploited in the past~\cite{Cui17,Koniusz17}. While promising, we believe that developing statistically-motivated pooling strategies for such higher-order information goes beyond the scope of this paper.


\section{Methodology}
In this section, we first introduce our second-order pooling strategy while explaining the statistical motivation behind it. We then provide an alternative interpretation of our approach yielding a lower complexity, study and display the empirical distributions of our network's representations, and finally discuss the relation of our model to the recent second-order pooling techniques.

\subsection{SMSO Pooling}
\label{subsec:smso}
Our goal is to design a general, parametric compression strategy for second-order deep networks. Furthermore, inspired by the fact that first-order deep networks inherently make use of Gaussian distributions for their feature representations, we reason about the statistical distribution of the network's intermediate representations so as to produce a final representation that is also Gaussian. Note that, while we introduce our SMSO pooling strategy within a CNN formalism, it applies to any method relying on second-order representations.

Formally, let $\mathbf{X} \in \mathbb{R}^{n\times c}$ be a data matrix, consisting of $n$ sample vectors of dimension $c$. For example, in the context of CNNs, $\bX$ contains the activations of the last convolutional layer, with $n = w \times h$ corresponding to the spatial resolution of the corresponding feature map. 
Here, we assume $\bx_i \in \mathbb{R}^c$ to follow a multivariate Gaussian distribution $\mathcal{N}_{c}({\bmu}, \mathbf{\Sigma})$. 
In practice, as discussed in~\cite{Goodfellow16,Ioffe15} and explained by the Central Limit Theorem, this can be achieved by using a \textit{linear} activation after the last convolutional layer, 
potentially followed by batch normalization~\cite{Ioffe15}.

\prg{Covariance Computation.}
Given the data matrix $\mathbf{X}$, traditional second-order pooling consists of computing a covariance matrix $\mathbf{Y} \in \mathbb{R}^{c\times c}$ as
\begin{equation}
\label{equ:cov}
	\mathbf{Y} = \frac{1}{n-1} \sum_{i=1}^{n} (\mathbf{x}_i - \bmu)(\mathbf{x}_i - \bmu)^T 
	= \frac{1}{n-1} \btX^T\btX\;,
\end{equation}
where $\btX$ denotes the mean-subtracted data matrix. 

The following definition, see, e.g.,~\cite{Johnson14}, determines the distribution of $\bY$. 
\begin{definition}
	If the elements $\mathbf{x}_i \in \mathbb{R}^{c}$ of a data matrix $\mathbf{X} \in \mathbb{R}^{n \times c}$ follow a zero mean multivariate Gaussian distribution $\bx_i \sim \mathcal{N}_{c}(0, \mathbf{\Sigma})$, then the covariance matrix $\mathbf{Y}$ of $\bX$ is said to follow a Wishart distribution, denoted by 
	\begin{equation}
	\mathbf{Y} = \mathbf{X}^T\mathbf{X} \sim W_{c}(\mathbf{\Sigma}, n)\;.
	\end{equation}
\end{definition}

Note that, in the bilinear CNN~\cite{Lin15}, the mean is typically not subtracted from the data. As such, the corresponding bilinear matrix follows a form of non-central Wishart distribution~\cite{James55}. 

\prg{Second-order Feature Compression.}
\label{sec:compress}
The standard way to use a second-order representation is to simply vectorize it~\cite{Lin15,Ionescu15}, potentially after some form of normalization~\cite{Lin17a,Li17a}. This, however, can yield very high-dimensional vectors that are cumbersome to deal with in practice. To avoid this, motivated by~\cite{Gao16,Kong17}, we propose to compress the second-order representation during vectorization. 
Here, we introduce a simple, yet effective, compression technique that, in contrast with~\cite{Gao16}, is parametric, and, as opposed to~\cite{Kong17}, amenable to general classifiers.

Specifically, we develop a parametric vectorization (PV) layer, which relies on trainable weights $\bW \in \mathbb{R}^{c\times p}$, with $p$ the dimension of the resulting vector. 
Each dimension $j$ of the vector $\bz$ output by this PV layer can be expressed as
\begin{equation}
\label{equ:pv}
	z_j = \mathbf{w}_j^T\mathbf{Y}\mathbf{w}_j\;,
\end{equation}
where $\bw_j$ is a column of $\bW$.

The distribution of each dimension $\bz_j$ is defined by the following theorem.
\begin{theorem}[Theorem 5.6 in~\cite{Johnson14}]
	\label{thm:chi-sq}
	If $\mathbf{Y} \in \mathbb{R}^{c \times c}$ follows a Wishart distribution $W_{c}(\mathbf{\Sigma}, n)$, and $\mathbf{w} \in \mathbb{R}^c$ and $\mathbf{w} \ne \mathbf{0}$, then
	\begin{equation}
	z = \frac{\mathbf{w}^T\mathbf{Yw}}{\mathbf{w}^T\mathbf{\Sigma w}} 
	\end{equation}
	follows a $\chi^2$ distribution with degree of freedom $n$, i.e., $z \sim \chi^2_n$.
\end{theorem}

From this theorem, we can see that each output dimension of our PV layer follows a scaled $\chi^2$ distribution $\gamma \chi^2_n$, where $\gamma = \mathbf{w}_j^T\mathbf{\Sigma}\bw_j$, with $\mathbf{\Sigma}$ the covariance matrix of the original multivariate Gaussian distribution. 

\prg{Transformation and normalization.} 
As shown above, each dimension of our current vector representation follows a $\chi^2$ distribution. However, as discussed above, first-order deep networks inherently exploit Gaussian distributions for their feature representations. To make our final representation also satisfy this property, we rely on the following theorem.

\begin{theorem}[\cite{Wilson31}]
\label{thm:sqrt}
	If $z \sim \chi^2_n$ with degree freedom $n$, then 
	\begin{equation}
		z' = \sqrt{2z} 
	\end{equation}
	converges to a Gaussian distribution with mean $\sqrt{2n-1}$ and standard deviation $\sigma = 1$ when $n$ is large, i.e., $z' \sim \mathcal{N}(\sqrt{2n-1},1)$.
\end{theorem}
Following this theorem, we therefore define our normalization as the transformation
\begin{equation}
	\mathbf{z}_j' = \sqrt{\alpha \mathbf{z}_j} - \sqrt{2n-1} \;,
\end{equation}
for each dimension $j$, where
we set $\alpha=2 / (\bw_j^T \mathbf{\Sigma} \bw_j)$ to correspond to Theorem~\ref{thm:sqrt}, while accounting for the factor $\gamma$ arising from our parametric vectorization above.
Note that other transformations, such as $\log(z)$ and $(z/n)^{1/3}$, are known to also converge to Gaussian distributions as $n$ increases~\cite{Bartlett46,Wilson31}. We show that these operations yield similar results to the one above in Section~\ref{subsec:config}.

Note that, according to Theorem~\ref{thm:sqrt}, the mean and variance of the resulting Gaussian distribution are determined by the degree of freedom $n$, which, in our case, corresponds to the number of samples used to compute the covariance matrix in Eq.~\ref{equ:cov}. Such pre-determined values, however, might limit the discriminative power of the resulting representation. To tackle this, we further rely on trainable scale and bias parameters, yielding a final representation
\begin{equation}
	\mathbf{z}_j'' = \beta_j + \gamma_j \mathbf{z}_j'\;,
	\label{eq:scale_bias}
\end{equation}
where $\gamma_j > 0, \beta_j \in \mathbb{R}$. 
Note that this transformation is also exploited by batch normalization. However, here, we do not need to compute the batch statistics during training, since Theorem~\ref{thm:sqrt} tells us that the batches follow a consistent distribution. 

Altogether, our SMSO pooling strategy, defined by the operations discussed above, 
yields a $p$-dimensional vector. This representation can then be passed to a classifier. 
It can easily be verified that the above-mentioned operations are differentiable, and the resulting deep network can thus be trained end-to-end.

\subsection{Alternative Computation}
\label{sec:alt}
Here, we derive an equivalent way to perform our SMSO pooling, with a lower complexity when $p$ is small, as shown in the supplementary material. Note, however, that our statistical reasoning is much clearer for the derivation of Section~\ref{subsec:smso} and was what motivated our approach.

To derive the alternative, we note that
\begin{align}
	\frac{1}{\sqrt{\alpha}}\mathbf{z}_j' &= \sqrt{\mathbf{w}_j^T\mathbf{Y}\mathbf{w}_j}  \\
	&= \sqrt{\mathbf{w}_j^T \left(\sum_{i=1}^{n} (\mathbf{x}_i - \mathbf{\mu})(\mathbf{x}_i - \mathbf{\mu})^T\right)\mathbf{w}_j} \\
	&= \sqrt{\sum_{i=1}^{n} \left(\mathbf{w}_j^T (\mathbf{x}_i - \mathbf{\mu})\right)\left((\mathbf{x}_i - \mathbf{\mu})^T\mathbf{w}_j\right)} \\
	&= \sqrt{\sum_{i=1}^{n} (\mathbf{w}_j^T\mathbf{\tilde{x}}_i)^2}\;, \label{eq:alt_final}
\end{align}

\noindent where $\mathbf{\tilde{x}}_i = \mathbf{x}_i - \mu$.

So, in essence, given $\bX$, $\bz'$ can be computed by performing a $1\times 1$ convolution, with weights shaped as $(1,1,c,p)$ and without bias, followed by a global $\ell_2$ pooling operation, and a scaling by the constant $\sqrt{\alpha}$. Note that $\ell_2$ pooling was introduced several years ago~\cite{Sermanet12}, but has been mostly ignored in the recent deep learning advances.  
By contrast, feature reduction with $1\times 1$ convolutions is widely utilized in first-order network designs~\cite{Szegedy15,He16}. In essence, this mathematically equivalent formulation shows that our second-order compression strategy can be achieved 
without explicitly computing
covariance matrices. Yet, our statistical analysis based on these covariance matrices remains valid.

\subsection{Relation to Other Methods}
\label{subsec:comparison}
In this section, we discuss the connections between our method and the other recent second-order pooling strategies in CNNs. In the supplementary material, we compare the computational complexity of different second-order methods with that of ours.

\prg{Normalization.}
Bilinear pooling (BP)~\cite{Lin15} also proposed to make use of a square-root as normalization operation. An important difference with our approach, however, is that BP directly vectorizes the matrix representation $\bY$. It is easy to see that the diagonal elements of $\bY$ follow a $\chi^2$ distribution, e.g., by taking $\bw$ in Theorem~\ref{thm:chi-sq} to be a vector with a single value 1 and the other values 0. Therefore, after normalization, some of the dimensions of the BP representation also follow a Gaussian distribution.
However, the off-diagonal elements follow a variance-gamma distribution, and, after square-root normalization, will not be Gaussian, thus making the different dimensions of the final representation follow inconsistent distributions.

In~\cite{Ionescu15} and~\cite{Li17a}, normalization was performed on the matrix $\bY$ directly, via a matrix logarithm and a matrix power normalization, respectively. As such, it is difficult to understand what distribution the elements of the final representation, obtained by standard vectorization, follow. 

\prg{Compression.}
The compact bilinear pooling (CBP) of~\cite{Gao16} exploits a compression scheme that has a form similar to ours in Eq.~\ref{equ:pv}. However, in~\cite{Gao16}, the projection vectors $\bw_j$ are random but fixed.
Making them trainable, as in our PV layer, increases the capacity of our model, and, as shown in Section~\ref{sec:experiments}, allows us to significantly outperform CBP. 

In~\cite{Kong17}, a model is developed specifically for a max-margin bilinear classifier. The parameter matrix of this classifier is approximated by a low-rank factorization, which translates to projecting the initial features to a lower-dimensional representation. As with our alternative formulation of Section~\ref{sec:alt}, the resulting bilinear classifier can be obtained without having to explicitly compute $\bY$. 
This classifier is formulated in terms of quantities of the form $\|\mathbf{U}^T \mathbf{X}_i\|^2_F$, where $\bU$ is a trainable low-rank weight matrix. 
In essence, this corresponds to removing the square-root operation from Eq.~\ref{eq:alt_final} and summing over all dimensions $j$.  
By contrast, our representation, ignoring the scale and bias of Eq.~\ref{eq:scale_bias}, is passed to a separate classification layer that computes a linear combination of the different dimensions with trainable weights, thus increasing the capacity of our model.

\begin{figure*}[t]
	\begin{center}
		\includegraphics[width=1\linewidth]{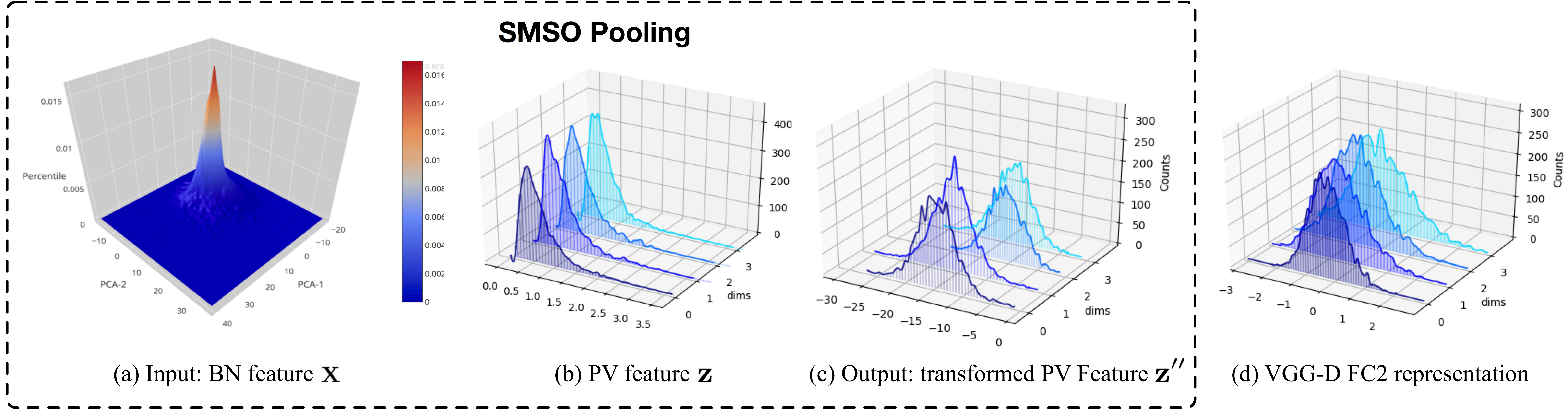}
	\end{center}
	\vspace{-0.3cm}
	\caption{{\bf Histograms of SMSO intermediate feature vectors.} We plot the distribution of (a) the initial features $\bX$, (b) the features after our PV layer $\bz$, (c) the final representation  $\bz''$ and, for comparison, (d) first-order features after the last fully-connected layer in VGG-D~\cite{Simonyan15}. Note that, as discussed in the text, these empirical distributions match the theoretical ones derived in Section~\ref{subsec:smso}, and our final representation does exploit the same type of distribution as first-order networks.
	}
	\label{fig:histogram-all}
\end{figure*}

\subsection{Empirical distributions of SMSO pooling}
\label{subsec:empirical_dist}
Our SMSO pooling strategy was motivated by considering the distribution of the representation at various stages in the network. Here, we study the empirical distributions of these features using the MINC dataset, discussed in Section~\ref{sec:experiments}, and with a model based on VGG-D. To this end, in Fig.~\ref{fig:histogram-all}, we provide a visualization of the distributions after the initial batch normalization (i.e., before computing the covariance matrix, see Section~\ref{subsec:implement} for details), after our PV layer, and after square-root transformation with trainable scaling and bias.
Specifically, for the initial features, because visualizing a Gaussian distribution in hundreds of dimensions is not feasible, we plot the distribution along the first 2 principal components. 
For our representations, where each dimension follows an independent Gaussian, we randomly select four dimensions and plot stacked histograms.
As expected from the theory, the initial features are close to Gaussian, and the features after our PV layer therefore follow a long-tailed $\chi^2$ distribution. The final features, after square-root normalization, scaling and bias, are much less skewed, and thus much closer to a Gaussian distribution, thus matching the type of distribution that the final representations of state-of-the-art deep networks follow, as shown in Fig.~\ref{fig:histogram-all}(d).
To further verify this, we conducted a Shapiro-Wilk test on the final representation. This resulted in a p-value of $0.19 > 0.05$, which means that the Gaussian assumption is not rejected, sustaining our claim.


\section{Experiments}
\label{sec:experiments}
Here, we first provide implementation details and introduce the baseline models. We then compare our approach to these baselines on four standard benchmark datasets, and provide an ablation study of our framework.
\subsection{Implementation Details}
\label{subsec:implement}
We evaluate our method on two popular network architectures: the VGG-D network of~\cite{Simonyan15} (a.k.a. VGG-16) and the ResNet-50 of~\cite{He16}. For all second-order models discussed below, i.e., ours and the baselines, we remove all the fully-connected layers and the last max pooling layer from VGG-D, that is, we truncate the model after the ReLU activation following \textit{conv5-3}. For ResNet-50, we remove the last global average pooling layer and take our initial features from the last residual block. As in~\cite{Li17a}, we add a $1\times 1$ convolutional layer to project the initial features to $c=256$ for all the experiments. Note that this is a linear layer, and thus makes the resulting features satisfy our Gaussian assumption.

Following common practice~\cite{Gao16,Kong17,Lin15,Li17a}, we rely on weights pre-trained on ImageNet and use stochastic gradient descent with an initial learning rate 10 times smaller than the one used to learn from scratch, i.e., 0.001 for VGG-D and 0.01 for ResNet-50. We then divide this weight by 10 when the validation loss has stopped decreasing for 8 epochs. We initialize the weights of the new layers, i.e., the $1 \times 1$ convolution, the PV layer and the classifier, with the strategy of~\cite{Glorot10}, i.e., random values drawn from a Gaussian distribution. We implemented our approach using Keras~\cite{chollet2015keras} with TensorFlow~\cite{Tensorflow15} as backend.

\subsection{Baseline Models}
\label{subsec:pooling}
We now describe the different baseline models that we compare our approach with. Note that the classifier is defined as a $k$-way softmax layer for all these models, as for ours, except for low-rank bilinear pooling, which was specifically designed to make use of a low-rank hinge loss. 

\prg{Original model:} This refers to the original, first-order, models, i.e., either VGG-D or ResNet-50, pre-trained on ImageNet and fine-tuned on the new data. Other than replacing the 1000-way ImageNet classification layer with a $k$-way one, we keep the original model settings described in~\cite{Simonyan15} and~\cite{He16}, respectively. 

\prg{Bilinear Pooling (BP)~\cite{Lin15}:} This corresponds to the original, uncompressed bilinear pooling strategy, with signed square-root and $\ell_2$ normalization after vanilla vectorization. In this case, we set $c=512$, as in the original paper, as the feature dimension before computing the second-order representation. If the original feature dimension does not match this value, i.e., with ResNet-50, we make use of an additional $1\times 1$ convolutional layer. Note that we observed that using either 512 or 256 as feature dimension made virtually no difference on the results. We therefore used $c=512$, which matches the original paper.

\prg{DeepO\textsubscript{2}P~\cite{Ionescu15}:} This refers to the original, uncompressed covariance-based model, with matrix logarithm and vanilla vectorization. Again, as in the original paper, we set $c=512$ as the feature dimension before computing the covariance matrix, by using an additional $1\times 1$ convolutional layer when necessary. 

\prg{Matrix Power Normalization (MPN)~\cite{Li17a}:} This model relies on a matrix square-root operation acting on the second-order representation. Following the original paper, we set $c=256$ by making use of an additional $1\times 1$ convolutional layer before second-order pooling. Note that the improved bilinear pooling of~\cite{Lin17a} has the same structure as MPN, and we do not report it as a separate baseline.

\prg{Compact bilinear pooling (CBP)~\cite{Gao16}: } We report the results of both versions of CBP: the Random Maclaurin (RM) one and the Tensor Sketch (TS) one. For both versions, we set the projection dimension to $d=8,192$, which was shown to achieve the same accuracy as BP, i.e., the best accuracy reported in~\cite{Gao16}. As in the original paper, we apply the same normalization as BP~\cite{Lin15}.

\prg{Low rank bilinear pooling (LRBP)~\cite{Kong17}:} This corresponds to the compression method dedicated to the bilinear SVM classifier. Following~\cite{Kong17}, we set the projection dimension to $m=100$ and its rank to $r=8$, and initialize the dimensionality reduction layer using the SVD of the Gram matrix computed from the entire validation set. Following the authors' implementation, we apply a scaled square-root with factor $2 \times 10^5$ after the \textit{conv5-3} ReLU, which seems to prevent the model from diverging. Furthermore, we found that training LRBP from the weights of BP fine-tuned on each dataset also helped convergence.

\subsection{Comparison to the Baselines} 

\begin{figure}[t]
	\begin{center}
		\includegraphics[width=1\linewidth]{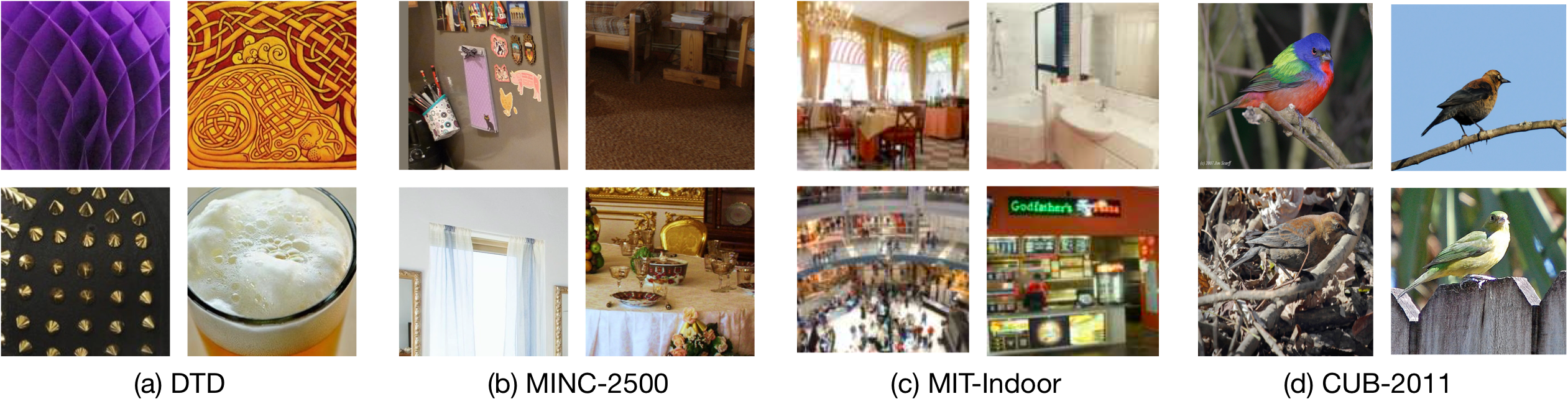}
	\end{center}
	\vspace{-0.7cm}
	\caption{Sample images from DTD, MINC-2500, MIT-Indoor and CUB.}
	\label{fig:dataset}
\end{figure}

Let us now compare the results of our model with those of the baselines described in Section~\ref{subsec:pooling}. To this end, we make use of four diverse benchmark image classification datasets, thus showing the versatility of our approach. These datasets are the Describing Texture Dataset (DTD)~\cite{Cimpoi14} for texture recognition, the challenging Material In Context (MINC-2500) dataset~\cite{Bell15} for large-scale material recognition in the wild, the MIT-Indoor dataset~\cite{Quattoni09} for indoor scene understanding and the Caltech-UCSD Bird (CUB) dataset~\cite{Wah2011} for fine-grained classification.
DTD contains 47 classes for a total of 5,640 images, mostly capturing the texture itself, with limited surrounding background. By contrast, MINC-2500, consisting of 57,500 images of 23 classes, depicts materials in their real-world environment, thus containing strong background information and making it more challenging. MIT-Indoor contains 15,620 images of 67 different indoor scenes, and, with DTD, has often been used to demonstrate the discriminative power of second-order representations. The CUB dataset contains 11,788 images of 200 different bird species.
 In Fig.~\ref{fig:dataset}, we provide a few samples from each dataset. For our experiments, we make use of the standard train-test splits released with these datasets. For DTD, MIT-Indoor and CUB, we define the input size as $448 \times 448$ for all the experiments. For the large-scale MINC-2500 dataset, we use $224 \times 224$ images for all models to speed up training. Note that a larger input size could potentially result in higher accuracies~\cite{Bell15}. For all datasets and all models, we use the same data augmentation strategy as in~\cite{Lin15,Lin17a}.

\begin{table}[t]
	\centering
	\begin{tabular}{l| c @{} rcrc @{} c @{}cccc}
	\toprule
	Model name & \phantomone & \multicolumn{1}{c}{Feature dim.} & & \multicolumn{1}{c}{\# param.}  & & \phantomtwo
	& DTD~\cite{Cimpoi14} 	&	MIT~\cite{Quattoni09}  & MINC~\cite{Bell15} & CUB~\cite{Wah2011} \\
	\cmidrule(r){1-1} 		\cmidrule(r){3-7} 		\cmidrule(r){8-11} 
		VGG-D~\cite{Simonyan15} 	&& 4,096 && 119.64M &&& 60.11 & 64.51 & 73.01 & 66.12\\
		BP~\cite{Lin15}  && $2.6\times 10^5$ 	&& 3.015M	&&& 67.50 & 77.55 & 74.50  & 81.02\\
		MPN~\cite{Li17a}	 && $32,896$ 			&& 0.752M	&&& 68.01 & 76.49 & 76.24 & 84.10\\
		DeepO$_2$P~\cite{Ionescu15} && $2.6\times 10^5$ && 3.015M &&& 66.07 & 72.35 & 69.29 & - \\
		\midrule
		CBP-TS~\cite{Gao16}  && 8,192 			&& 0.189M	&&& 67.71 & 76.83 & 73.28	 & 84.00 \\ 
		CBP-RM~\cite{Gao16}  && 8,192 			&& 0.189M	&&& 63.24 & 73.89 & 73.54  & 83.86 \\ 
		LRBP~\cite{Kong17}   && 100 			&& 0.068M	&&& 65.80 & 73.59 & 69.10  & 84.21 \\
		\midrule
		SMSO~(\textit{Ours}) && 64				&& 0.013M  &&& 68.18	& 75.37	& 74.18	 & 82.66 \\
		SMSO~(\textit{Ours}) && 2,048	 		&& 0.057M  &&& \textbf{69.26}	& \textbf{79.45} &  \textbf{78.00} & \textbf{85.01} \\
		\bottomrule
	\end{tabular}
	\vspace{+0.05cm}
	\caption{{\bf Comparison of VGG-D based models.} We report the top 1 classification accuracy (in \%) of the original VGG-D model, uncompressed second-order models with different normalization strategies (BP, DeepO$_2$P, MPN), second-order compression methods (CBP-TS, CBP-RM, LRBP), and our approach (SMSO) with different PV dimensions. Note that our approach significantly outperforms all the baselines despite a more compact final representation (Feature dim.) and much fewer parameters (\# param is the number of trainable parameters after the last convolutional layer). }
	\label{tab:vgg}
\end{table}

\begin{figure}[t]
	\begin{center}
		\includegraphics[width=1\linewidth]{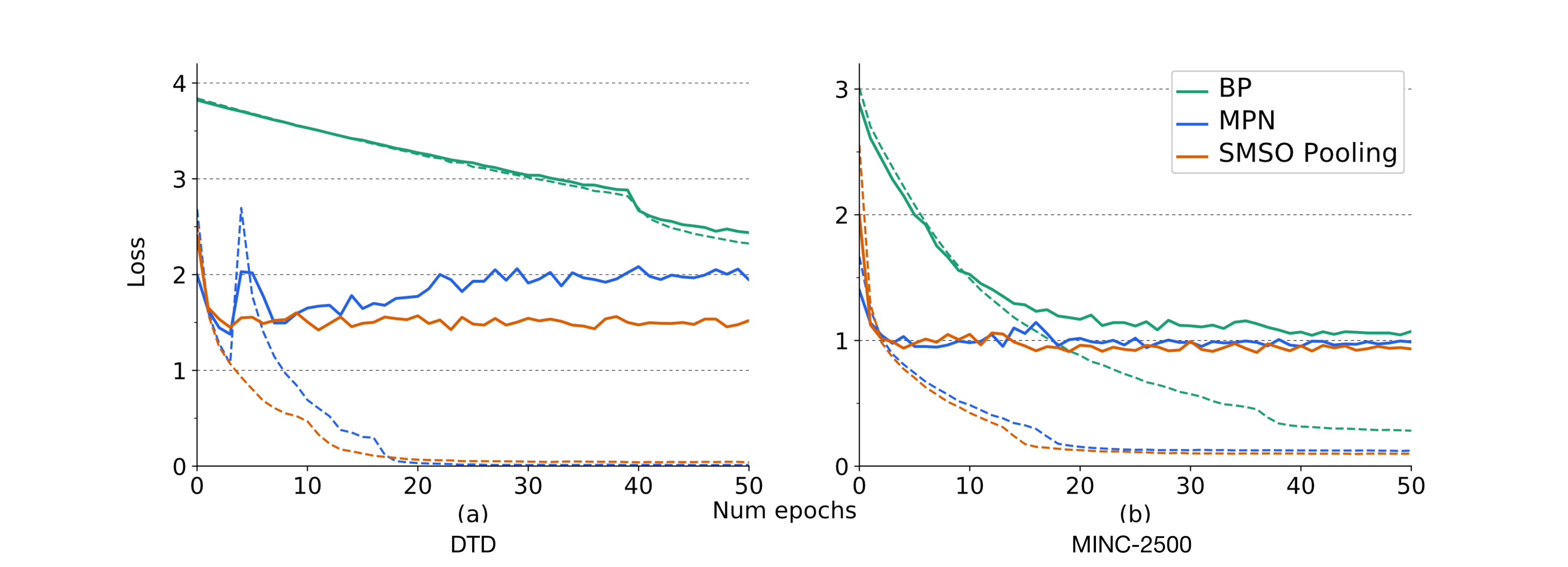}
	\end{center}
	\vspace{-0.7cm}
	\caption{{\bf Training and validation loss curves.} We plot the training (dashed) and validation (solid) loss values as a function of the number of training epochs for our SMSO pooling strategy~(orange), BP~(green) and MPN~(blue) on DTD~(a) and MINC-2500~(b). Our models clearly converge faster than BP, and tend to be more stable than MPN, particularly on the smaller-scale DTD dataset.
	}
	\label{fig:converge}
\end{figure}

\prg{Experiments with VGG-D.}
We first discuss the results obtained with the VGG-D architecture as base model. These results are reported in Table~\ref{tab:vgg} for all models and all datasets. In short, our SMSO framework with PV dimension $p=2,048$ outperforms all the baselines by a significant margin on all three datasets. In particular, our accuracy is $7\%$ to $19\%$ higher than the original VGG-D, 
with \emph{much} fewer parameters,
thus showing the benefits of exploiting second-order features. MPN is the best-performing baseline, but, besides the fact that we consistently outperform it, has a much higher computational complexity and run time, as shown in the supplementary material.
The second-order compression methods (CBP and LRBP) underperform the uncompressed models on average. By contrast, even with $p=64$, we outperform most baselines, with a model that corresponds to 10\% of the parameters of the most compact baseline.

In Fig.~\ref{fig:converge}, we compare the training and validation loss curves of our approach with those of the best-performing baselines, BP and MPN, on DTD and MINC. Note that our model converges much faster than BP and tends to be more stable than MPN, particularly on DTD. This, we believe, is due to the fact that we rely on basic algebraic operations, instead of the eigenvalue decomposition involved in MPN whose gradient can be difficult to compute, particularly in the presence of small or equal eigenvalues~\cite{Ionescu15}.

During these VGG-D based experiments, we have observed that, in practice, LRBP was difficult to train, being very sensitive to the learning rate, which we had to manually adapt throughout training. Because of this, and the fact that LRBP yields lower accuracy than uncompressed models, we do not include this baseline in the remaining experiments. We also exclude DeepO$_2$P from the next experiments, because of its consistently lower accuracy.

\begin{table*}[t]
	\centering

	\begin{tabular}{l| c @{} rcrc @{} c @{}cccc}
		\toprule
		Model name & \phantomone & \multicolumn{1}{c}{Feature dim.} & & \multicolumn{1}{c}{\# param.}  & & \phantomtwo
		& DTD~\cite{Cimpoi14} 	&	MIT~\cite{Quattoni09}  & MINC~\cite{Bell15} & CUB~\cite{Wah2011} \\
		\cmidrule(r){1-1} 		\cmidrule(r){3-7} 		\cmidrule(r){8-11} 
		ResNet-50~\cite{He16} &	& 2,048 & & 4K  &	& & 71.45 & 76.45 & 79.12 & 74.51 \\
		BP~\cite{Lin15} & & $32,896$ 	&& 752K &	&& 69.37 & 68.35 & 79.05  & 82.70 \\
		MPN~\cite{Li17a}&& $32,896$ 	&	& 752K & && 71.10 & 72.12 & 79.83 & 85.43 \\
		\midrule
		CBP-TS~\cite{Gao16} &&  8,192 	&& 189K & && 65.30 & 72.60 & 75.91	 & 77.35 \\ 
		CBP-RM~\cite{Gao16} && 8,192 	&& 189K & && 62.35 & 67.81 & 74.15  & - \\ 
		\midrule
		SMSO~(\textit{Ours})  &	& 64	&& 13K & 	&& 71.03 & 76.31 & 79.17	 & 81.98 \\
		SMSO~(\textit{Ours})  &	& 2,048 && 57K &	&& \textbf{72.51}	& \textbf{79.68} &  \textbf{81.33} & \textbf{85.77} \\
		\bottomrule
	\end{tabular}
	\vspace{+0.05cm}
	\caption{{\bf Comparison of ResNet-50 based models.} We report the top 1 classification accuracy (in \%) of the original ResNet-50 model, uncompressed second-order models with different normalization strategies (BP, MPN), second-order compression methods (CBP-TS, CBP-RM), and our approach (SMSO). Note that, as in the VGG-D case, our model outperforms all the baselines, including the original ResNet-50, which is not the case of most second-order baselines. It also yields much more compact models than the second-order baselines. (\# params. refers to the same quantity as in Table 1.)}
	\label{tab:resnet}

\end{table*} 

\begin{figure}[t]

	\begin{center}
		\includegraphics[width=0.9\linewidth]{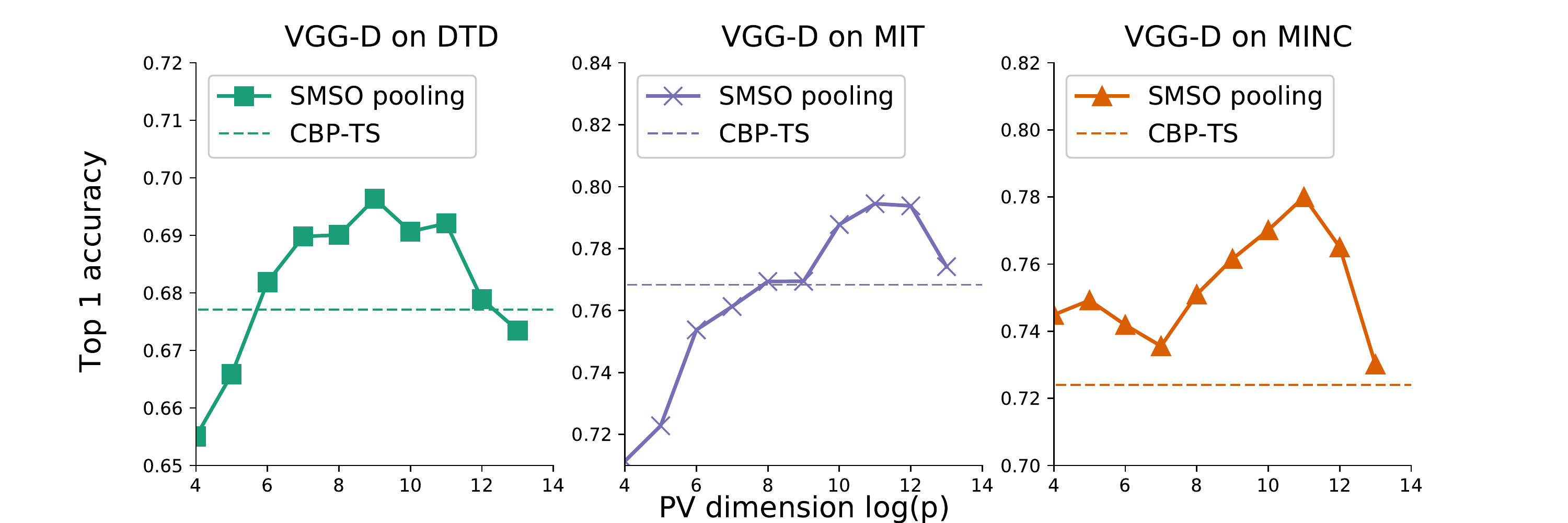}
	\end{center}
	\vspace{-0.7cm}
	\caption{{\bf Influence of the PV dimension $p$.} We plot the top 1 accuracy as a function of the value $p$ in logarithmic scale on MIT (left), MINC (middle) and DTD (right). Note that accuracy is quite stable over large ranges of $p$ values, yielding as good results as the best-performing compression baseline (CBP-TS) with as few as $p=64$ dimensions, corresponding to only 10\% of the parameters of CBP-TS.}
	\label{fig:pv_cross}

\end{figure}

\prg{Experiments with ResNet-50.} 
To further show the generality of our approach, we make use of the more recent, very deep ResNet-50~\cite{He16} architecture as base network. Table~\ref{tab:resnet} provides the results of our SMSO framework with $p=64$ and $p=2,048$, and of the baselines. 
In essence, the conclusions remain unchanged; we outperform all the baselines for $p=2,048$. Note that, here, however, the second-order baselines typically do not even outperform the original ResNet-50, whose results are significantly higher than the VGG-D ones. By contrast, our model is able to leverage this improvement of the base model and to further increase its accuracy by appropriately exploiting second-order features. 

\subsection{Ablation Study}
\label{subsec:config}
We evaluate the influence of different components of our model on our results. 
  
\prg{Influence of the PV dimension.} 
In our experiments, we proposed to set $p=2,048$ or $p=64$.
We now investigate the influence of this parameter on our results. To this end, we vary $p$ in the range $[2^4,2^{13}]$ by steps corresponding to a factor 2. The curves for this experiment on the validation data of the three datasets with VGG-D based models are provided in Fig.~\ref{fig:pv_cross}. Note that our model is quite robust to the exact value of this parameter, with stable accuracies outperforming the best compression baseline for each dataset over large ranges. More importantly, even with $p=64$, our model yields as good results as the best compression method, CBP-TS, with only $10\%$ of its parameters.

\prg{Comparison of different distributions and transformations.} 
We conduct experiments to compare different final feature distributions
on MINC-2500 with a VGG-D based model. The results are provided in Table~\ref{tab:distributions}.
Without our PV compression and without transformation or normalization, the resulting features follow a Wishart distribution, yielding an accuracy of 75.97\%, which is comparable to BP~\cite{Lin15}. Adding our PV layer $p=2,048$, but not using any transformation or normalization, yields $\chi^2$-distributed features and an accuracy similar to the previous one. This suggests that our parametric compression is effective, since we obtain similar accuracy with much fewer parameters. Including the square-root transformation, but without the additional scale and bias of Eq.~\ref{eq:scale_bias}, increases the accuracy to $76.32\%$. Additionally learning the scale and bias boosts the accuracy to $78.00\%$, thus showing empirically the benefits of Gaussian-distributed features over other distributions.

In the last two columns of Table~\ref{tab:distributions}, we report the results of different transformations that bring the $\chi^2$-distributed features to a Gaussian distribution, i.e., the cubic-root and the element-wise logarithm. Note that these two transformations yield accuracies similar to those obtained with the square-root. More importantly, all transformations yield higher accuracies than not using any ($76.14\%$), which further evidences the benefits of Gaussian-distributed features. 

\begin{table}[t]
	\centering
	\small
	\begin{tabular}{l| c @{} l cccccccc @{} c }
		\toprule
		Vec. & Flatten & \phantomone & \multicolumn{5}{c}{PV} & & \multicolumn{3}{c}{PV} \\
		\cmidrule(r){1-1} \cmidrule(r){2-3} \cmidrule(r){4-8} \cmidrule(r){9-12}
		Trans.	& - 	&		& - 	& Sqrt 	& Sqrt 	& Sqrt	& -		 & \phantomone & Sqrt	& Log & $\sqrt[3]{\phantomone}$ 	\\ 
		Norm.	& - 	&		& - 	& - 	&$\beta$&$\gamma$ & $\beta, \gamma$ 
		& \phantomone
		&\multicolumn{3}{c}{$\beta, \gamma$} 	\\ 
		\cmidrule(r){1-1} \cmidrule(r){2-3} \cmidrule(r){4-8} \cmidrule(r){9-12}
		Dist.	& $W_n(\Sigma)$ & & $\chi^2_n$  & \phantomone $\mathcal{N}(\mu,1)$ \phantomone & $\mathcal{N}(\beta, 1)$ 
		& \phantomone $\mathcal{N}(\mu,\gamma^2)$ \phantomone &  $\gamma \chi^2_n + \beta$ 
		& \phantomone
		& \multicolumn{3}{c}{$\mathcal{N}(\beta,\gamma^2)$} \\
		Acc.	& 75.97		&	& 75.32	& 76.32	& 77.12	& 76.47 & 76.14 & \phantomone & 78.00	&\phantomone 77.86 \phantomone  & 77.17 \\ 
		
		\bottomrule
	\end{tabular}
	\vspace{0.1cm}
	\caption{{\bf Comparison of different final feature distributions.} 
	We report the results of different combinations of vectorization (vec.), transformation (trans.) and normalization (norm.) strategies, yielding different final feature distributions. Here, $\mu = \sqrt{2n-1}$ from Theorem~\ref{thm:sqrt}. Ultimately, these results show that bringing the data back to a Gaussian distribution with a trainable scale and bias yields higher accuracies. 
}
\label{tab:distributions}
\end{table}


\section{Conclusion}
We have introduced a general and parametric compression strategy for second-order deep networks, motivated by a statistical analysis of the distribution of the network's intermediate representations.
Our SMSO pooling strategy outperforms the state-of-the-art first-order and second-order models, with higher accuracies than other compression techniques for up to 90\% parameter reduction.
With a ResNet-50 base architecture, it is the only second-order model to consistently outperform the original one. 
While Gaussian distributions have proven effective here and for first-order models, there is no guarantee that they are truly optimal.
In the future, we will study if other transformations yielding non-Gaussian distributions can help further improve second-order models.

	%
	%
	%
	%
		
\end{document}